\documentclass[lettersize,journal]{IEEEtran}
\usepackage{times}  
\usepackage{helvet}  
\usepackage{courier}  
\usepackage[hyphens]{url}  
\usepackage{graphicx} 
\usepackage{natbib}  
\usepackage{caption} 
\usepackage{algorithm}
\usepackage{algorithmic}
\usepackage{booktabs}
\usepackage{multirow}
\usepackage{comment}
\usepackage{xcolor}
\usepackage{subcaption}
\usepackage{hyperref}
\usepackage[accsupp]{axessibility}  
\usepackage{newfloat}
\usepackage{listings}
\DeclareCaptionStyle{ruled}{labelfont=normalfont,labelsep=colon,strut=off} 
\floatstyle{ruled}
\newfloat{listing}{tb}{lst}{}
\floatname{listing}{Listing}
\title{ED$^4$: Explicit Data-level Debiasing for Deepfake Detection}
\author{
    Jikang Cheng, Ying Zhang, Qin Zou, \textit{Senior Member, IEEE}, Zhiyuan Yan, Chao Liang, \textit{Member, IEEE}, Zhongyuan Wang, \textit{Member, IEEE}, Chen Li  
    \thanks{Jikang Cheng, Chao Liang, Qin Zou, and Zhongyuan Wang are with the School of computer science, Wuhan University, Wuhan 430000, China. (e-mail: \{chengjikang, cliang, qzou\}@whu.edu.cn, wzy\_hope@163.com). 

    Ying Zhang and Chen Li are with WeChat, Tencent Inc., Shenzhen, China. (e-mail: \{yinggzhang, chaselli\}@Tencent.cn).

    Zhiyuan Yan is with the School of electronic and computer engineering, Peking University Shenzhen Graduate School. 
    
    Work done during Jikang Cheng was an intern at WeChat.

    Our work was supported by the National Natural Science Foundation of China (62171324, 62371350, 62372339);
    Key Science and Technology Research Project of Xinjiang Production and Construction Corps in 2025. Corresponding author: Zhongyuan Wang.
 }
}

\usepackage{bibentry}

\newcommand{\etal}{\textit{et al.}}
\definecolor{myBrown}{rgb}{0.769, 0.349, 0.067}
\definecolor{myBlue}{rgb}{0.357, 0.608, 0.835}

\newcommand{\reT}[1]{#1}
\newcommand{\reG}[1]{#1}

\begin{document}

\maketitle

\begin{abstract}
%

Learning intrinsic bias from limited data has been considered the main reason for the failure of deepfake detection with generalizability. Apart from the discovered content and specific-forgery bias, we reveal a novel \textbf{spatial bias}, where detectors inertly anticipate observing structural forgery clues appearing at the image center, also can lead to the poor generalization of existing methods. We present ED$^4$, a simple and effective strategy, to address aforementioned biases \textit{explicitly at the data level} in a unified framework rather than implicit disentanglement via network design. In particular, we develop ClockMix to produce facial structure preserved mixtures with arbitrary samples, which allows the detector to learn from an exponentially extended data distribution with much more diverse identities, backgrounds, local manipulation traces, and the co-occurrence of multiple forgery artifacts. We further propose the Adversarial Spatial Consistency Module (AdvSCM) to prevent extracting features with spatial bias, which adversarially generates spatial-inconsistent images and constrains their extracted feature to be consistent.
As a model-agnostic debiasing strategy, ED$^4$ is \textbf{plug-and-play}: it can be integrated with various deepfake detectors to obtain significant benefits. 
We conduct extensive experiments to demonstrate its effectiveness and superiority over existing deepfake detection approaches. Code is available at \textit{\href{https://github.com/beautyremain/ED4}{https://github.com/beautyremain/ED4}}.

\end{abstract}

\section{Introduction}
The growing threats posed by deepfake technology in social media have heightened the necessity of detecting malicious deepfake content. Hence, deepfake detection technology attracts increasing attention from the research community. Most deepfake detectors perform promisingly when dealing with in-dataset images. However, their effectiveness faces significant challenges when transferred to unseen data distributions, 
which restricts their ability in practical usage.

\begin{figure}[ht]
     \centering
     \includegraphics[width=0.98\linewidth]{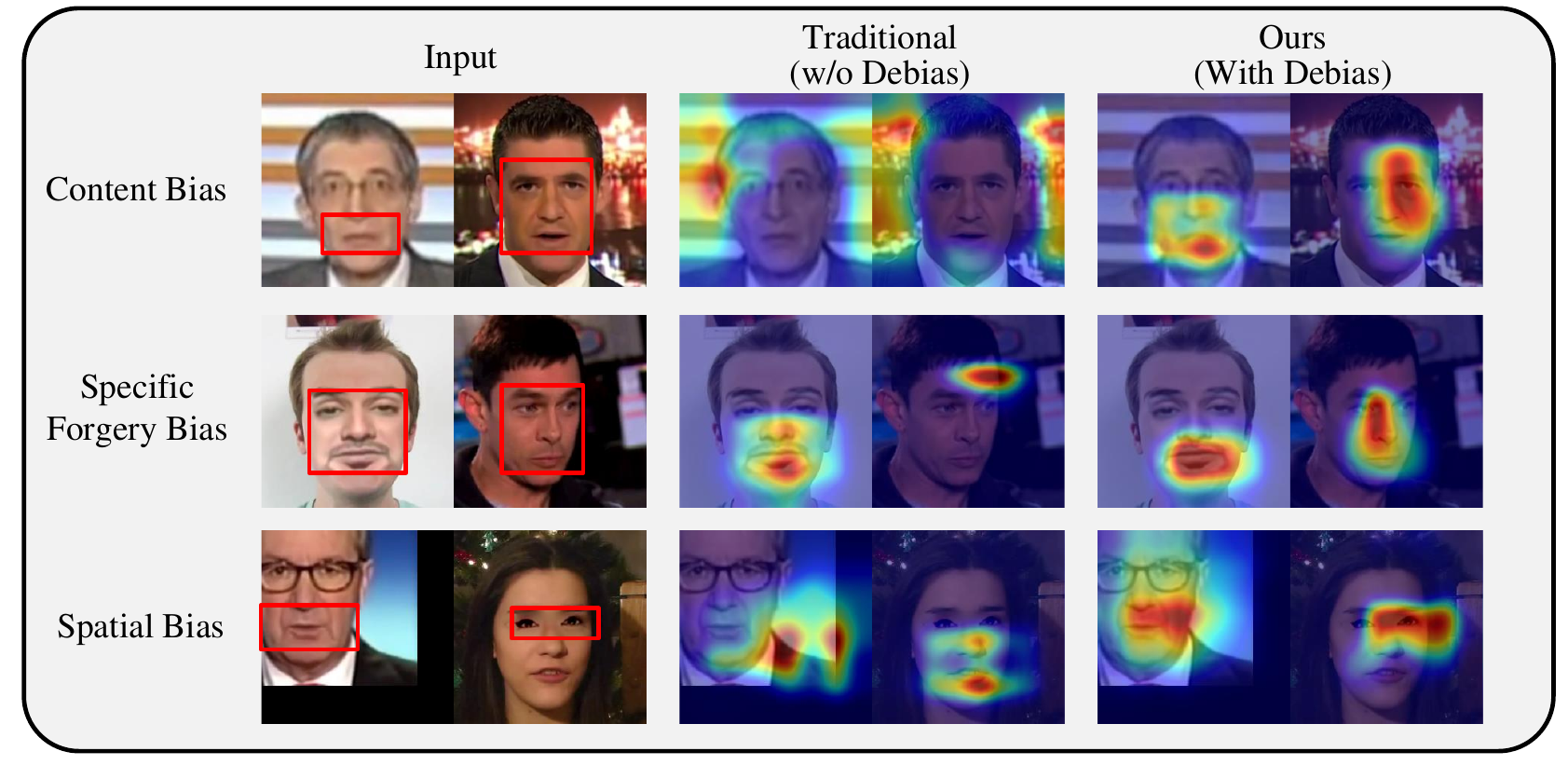}
     \caption{Illustration of different biases in deepfake detection.  Two inputs for specific-forgery bias are in-dataset and cross-dataset, respectively.
     The red rectangle indicates the region containing the ground-truth forgery traces. We discover that the traditional detector may mistakenly focus on
     \textbf{\textit{Content Bias}}: The striking background without considering the possible forgery artifacts presented in the faces.
     \textbf{\textit{ Specific-Forgery Bias}}: Specific in-dataset artifacts.
    \textbf{\textit{Spatial Bias}}: 
     Structural forgery clues at the image center despite the faces being shifted or locally manipulated.}
     \label{fig:bias}
 \end{figure}
 

To improve the generalization ability, existing deepfake detectors make attempts from multiple aspects \cite{capsule,lips,huang2023implicit,Disentangle,ucf,tip2,tip3}. Among them, methods targeted at removing the model bias 
hold the view that, deepfake detectors could easily learn \reG{biased} information to categorize input data, rather than digging the intrinsic forgery evidence. 
Liang \etal \cite{Disentangle} demonstrate that detectors erroneously learn the identity and background information, which can be referred to as content bias. Then, they design an encoder-decoder network, attempting to achieve feature disentanglement through implicit network constraints. 
Yan \etal \cite{ucf} contend that detectors tend to focus on the forgery artifacts related to one specific manipulation method, thereby overlooking common forgery artifacts. To address such specific-forgery bias, they also adopt a similar encoder-decoder network to disentangle common forgery features. 
With inspiring observations and analyses, these methods struggled to \textbf{implicitly} separate the generalizable forgery features from bias with an encoder-decoder network, which is \textit{non-intuitive and insufficient to guarantee successful disentanglement}. 
Broadly speaking, the fake synthesis methods can be considered as a way of removing the model bias by creating new fake images. However, their performance is hindered by the limitation in mixing identities \cite{SBI,leakage}, fake regions \cite{adv}, or blend types \cite{xray}. 


Apart from the \textit{content bias} and \textit{specific-forgery bias}, as shown in Fig.~\ref{fig:bias}, we first reveal a novel model bias that has been previously underestimated, which we term \textit{spatial bias}. Specifically, we notice that detectors usually expect to observe structural forgery clues at the image center, regardless of the actual facial location or the existence of local forgery artifacts. Therefore, the learned forgery information during training is oversimplified and the model sensitivity to forgery artifacts is severely compromised. Moreover, spatial bias can significantly undermine the robustness of the detector by posing challenges against spatial deviation, which commonly occurs in face detection and preprocessing.  


In this paper, we propose Explicit Data-level Debiasing for Deepfake Detection (ED$^4$), aiming to address the aforementioned three biases in a unified framework.  Specifically, we introduce two effective modules to remove the model bias in a hybrid manner: ClockMix, and Adversarial Spatial Consistency Module (AdvSCM).
ClockMix aims to address the content and specific-forgery biases via the clockwise mixing of different images. It takes multiple images with arbitrary faces as input, and then performs sector-based mosaicing centered on the face, to obtain a mixture containing different backgrounds, identities, and forgery artifacts. Different from existing mixing strategies, ClockMix is carefully designed for face forgery detection with high flexibility and surprising effectiveness. 
Then, we propose AdvSCM to tackle spatial bias. Specifically, we introduce an adversarial generator to produce shuffled images with larger spatial inconsistency. Subsequently, we enforce the detector to learn consistent forgery feature representations of images with distinct spatially inconsistent versions. 
By introducing AdvSCM, the deepfake detector is forced to ignore spatial deviations and pay more attention to informative local regions, thus gaining stronger discriminant power in identifying subtle forgery patterns rather than inertly anticipating the observation of structural forgery clues at image centers.


\begin{figure}[t]
    \centering
    \includegraphics[width=0.9\linewidth]{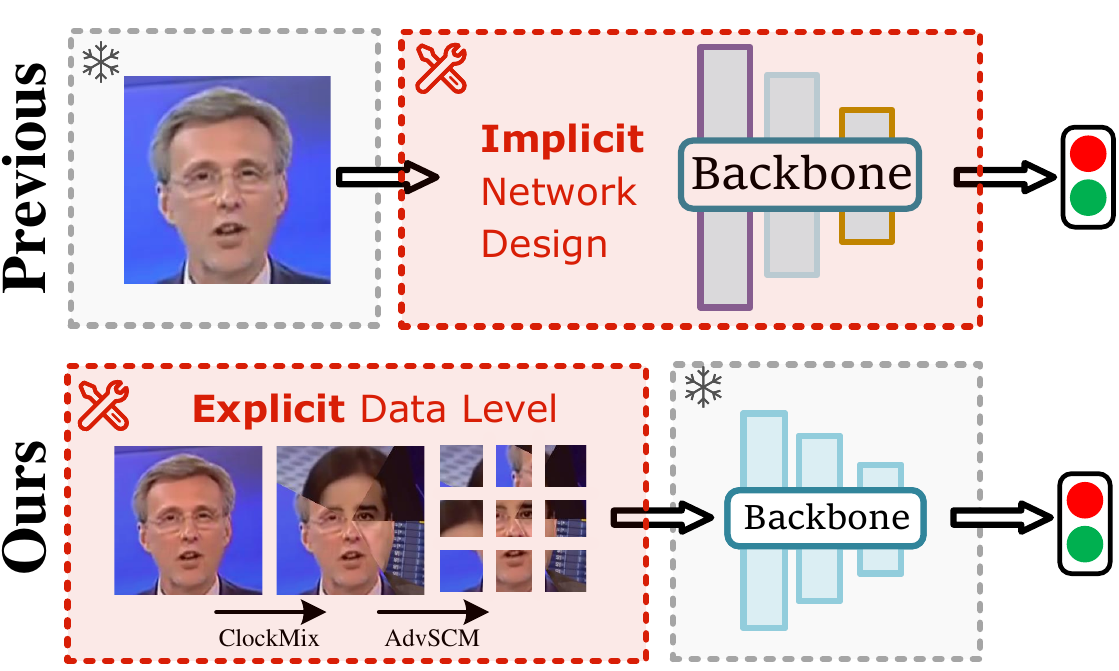}
    \caption{\reT{Comparisons between debiasing via implicit network design and explicit data level. The red regions indicate the target part for algorithms' modifications, whereas the parts in the gray regions remain unaltered by the algorithms.  Note that the term ``unaltered backbone'' refers exclusively to the network architecture, while its parameters are typically optimized during training.}}
    \label{fig:pipe-compare}
\end{figure}
\reT{As shown in Fig.~\ref{fig:pipe-compare}, the explicit debiasing method pre-processes the training samples instead of altering the backbone network architecture, allowing plug-and-play effectiveness, more intuitive visual understanding, and easier real-world implementation.}
Therefore, by explicitly removing the model bias at the data level, ED$^4$ is a superior alternative to the implicit disentanglement achieved by designing networks~\cite{Disentangle,ucf}, as well as to traditional fake synthesis approaches~\cite{adv,leakage,xray,SBI}.  The advantages of ED$^4$ could be summarized as:
\begin{itemize}
    \item ED$^4$ feeds the network with images covering more varied identities, backgrounds, forgery patterns, and spatial distributions, which can intuitively improve the network generalization with the \textbf{increased data diversity}. 
    \item ED$^4$ achieves stronger data augmentation by mosaicing arbitrary face images, \textbf{with no limitation} in image labels, paired identities, or fake regions.
    \item ED$^4$ is model-agnostic and can be easily applied as a \textbf{plug-and-play} module to improve the deepfake detection methods, without introducing computation overhead during inference.
\end{itemize}

\section{Related Works}
\subsection{Deepfake Detection}
Deepfake detection aims to classify the authority of an input image, which may be forged by deepfake manipulations.
Some approaches focus primarily on specific facial representations, such as lip movement \cite{lips} and action unit consistency \cite{aunet}. Meanwhile, various studies are dedicated to developing optimal neural network structures to improve detection performance, such as MesoNet~\cite{mesonet}, Xception \cite{xception}, and CapsuleNet \cite{capsule}. In the frequency domain, SPSL \cite{spsl} and SRM \cite{srm} introduce the phase spectrum and high-frequency noises to enhance the forgery information for training. IFFD~\cite{tip1} learns patch-channel correspondence to achieve a more interpretable deepfake detection. These methods focus on identifying specific vulnerabilities inherent in Deepfake methods, achieving significant success in detecting fake images that exhibit these specific vulnerabilities.

Owing to the limited data variations, certain specific data characteristics 
may be exclusively found within the data associated with a particular label. Subsequently, the detector may take a shortcut by solely learning the correlation between certain characteristics and the label, neglecting to achieve an in-depth comprehension of forgery features. We refer to this inert mapping as model bias. 
Dong \etal \cite{leakage} and Huang \etal \cite{huang2023implicit} posit that the target face identity used during face swapping remains in the fake face, leading to implicit identity leakage. Liang \etal \cite{Disentangle} design an encoder-decoder network to implicitly disentangle the content bias, attempting to obtain a forgery-only feature for detection. 
UCF \cite{ucf} utilizes indirect constraints and network design to implicitly disentangle common forgery features from specific features, guiding the learning toward generalized forgery information. 

While these methods demonstrate the significance of content and specific-forgery biases, the implicit disentanglement achieved by the encoder-decoder network design appeared overly complex and indirect. Namely, the components for decoding are redundant during inference and are limited to indirectly constraining the encoder for the disentanglement during training.


\subsection{Data Augmentation and Synthesis for Deepfake Detection}
\begin{figure*}[ht]
    \centering
    \includegraphics[width=0.95\linewidth]{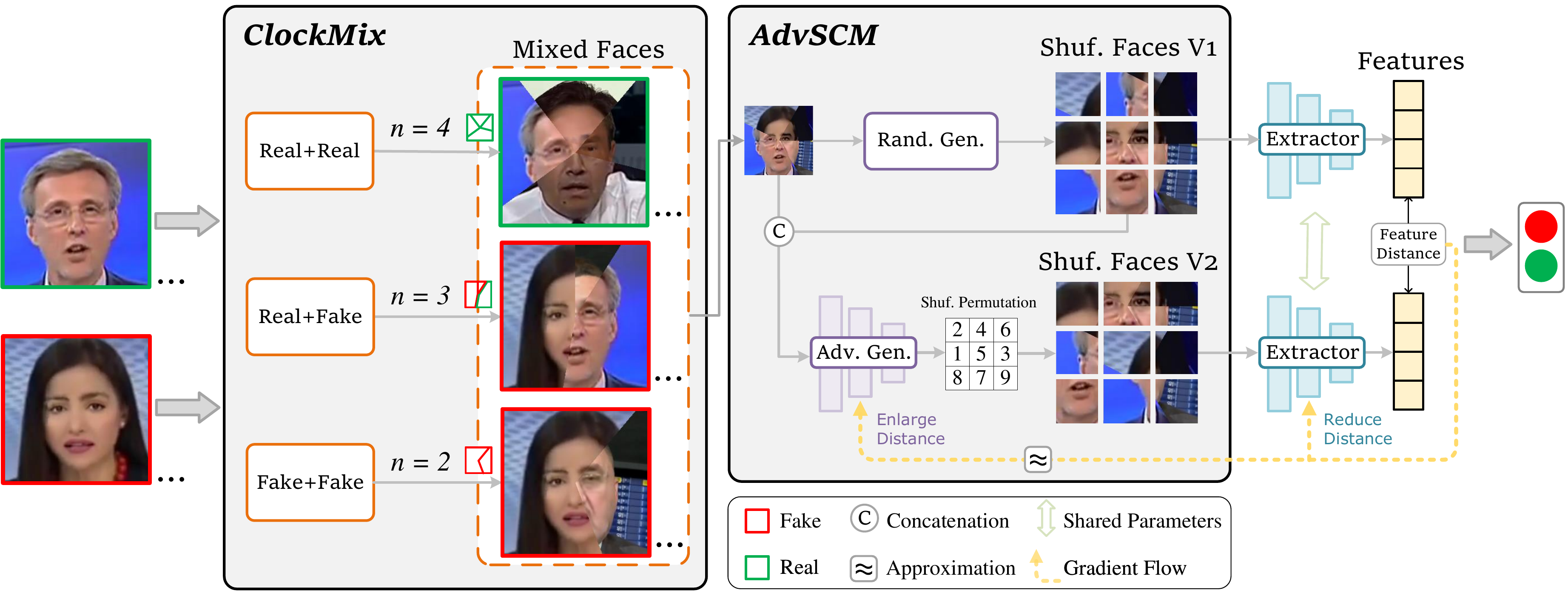}
    \caption{Overall framework of the proposed approach. \reT{Both real and fake training samples are first processed by ClockMix, generating multiple mixed samples according to the mixing combinations. Then, AdvSCM is applied to the mixed samples to remove spatial bias in an adversarial manner. \textit{Concatenation} represents combining two groups of images together as the new input. \textit{Approximation} indicates the non-differentiable \textit{gradient flow} for network optimization is approximated via the Reinforce algorithm, instead of using direct back-propagation. Finally, the ED$^4$ processed samples are input to the backbone detector for supervised learning.} }
    \label{fig:main_arch}
\end{figure*}
Data augmentation is widely employed during the training of neural networks to enhance the generalizability of the model, which enriches data diversity by implementing spatial transformations to the training samples. Various traditional data augmentation methods have been demonstrated to exhibit significant impacts on the generalizability of deepfake detectors. For instance, Liang \etal \cite{Disentangle} attempts to incorporate the feature-level Mixup, while Wang \etal \cite{masking} designed a specific version of random erasing \cite{erasing} based on attention for deepfake detection. DCL~\cite{dcl} proposes to leverage groups of augmented images via dual contrastive learning.
\reT{TALL~\cite{tall} incorporates consecutive frames in the same video and re-arranges their layout in a thumbnail manner. By doing this, it can calculate self-attention among local patches from different adjacent frames. However, TALL cannot address content bias since frames in the same video share the same content.}
Additionally, many strategies introduce diverse data synthesis approaches that are tailored to the deepfake detector. Unlike traditional augmentation, these synthesis methods generate new data from the existing training data and with more defined objectives. Specifically, Face X-ray \cite{xray} simulates blending artifacts by replacing one pristine face with another pristine face with the nearest landmarks. SLADD \cite{adv} adopts an adversarial training strategy to select harder forgery artifacts and blend them to pristine faces, obtaining more challenging samples to enhance the sensitivity of the detector. FWA \cite{FWA} and SBI \cite{SBI} simulated the quality inconsistency by replacing a pristine face with a transformed version of itself, forcing detectors to heighten sensitivity to forged information.

The existing data synthesis methods \reG{mix} two paired images with specific regions, which limits the diversity of data mixing. In contrast, our method employs a comprehensive random integration of multiple arbitrary faces with any label, which is beneficial to bias removal.
\section{Analysis of Model Bias}
\subsection{Model Bias and Spurious Correlation}
Model bias is a common issue in deep learning and has been widely studied~\cite{butterfly,ood-survey,tench,domainnet}. It is not inherently linked to the network architecture of the model (\textit{e.g.}, CNN, RNN, Transformer), but rather arises from biases present in the training data distribution. Specifically, the network learns spurious correlations between the input samples and the output targets from the data. For instance, most images labeled as ``butterfly" in ImageNet also contain flowers~\cite{butterfly}, and most images labeled as ``tench" depict a fisherman holding the fish~\cite{tench}. Consequently, the biases present in these data distributions lead the model to inertly establish spurious correlations between certain labels and features that are irrelevant to the task objective~\cite{sc1}. Finally, the bias makes the networks suffer from \textit{Lack of robustness}, \textit{Lack of reliable confidence estimates}, and \textit{Suboptimal generalization}~\cite{simple-bias}.
\subsection{In the Context of Deepfake Detection}
Here, we illustrate the influence of model bias in \textit{deepfake detection}. As shown in the first row of Fig.~\ref{fig:bias}, if the training data \reG{only } contains images labeled as ``fake" with striped backgrounds or prominent flashes, the model will tend to erroneously associate \reG{those patterns} with the "fake" label. This not only causes the detector to make misclassifications due to learned spurious correlations based on certain features, but also reduces the detector's sensitivity to core forgery clues, thereby further degrading its performance. In this paper, we address three major biases in deepfake detection, that is, content bias, specific forgery bias, and spatial bias.\\
\textbf{Content and Specific Forgery Biases.}
Content and specific forgery biases are introduced by \cite{Disentangle} and \cite{ucf} to deepfake detection. Specifically, Liang \etal recognize the issue of content bias in deepfake detection, suggesting that networks might mistakenly treat certain identities or backgrounds as either Fake or Real based on learned biases from the training data without learning the actual forgery information. Yan \etal define specific forgery bias as the network overly focusing on forgery clues that only exist in specific forgery techniques in training data, thereby undermining its generalization ability to unseen forgeries.\\
\textbf{The Novel Spatial Bias.}
In this paper, besides content and specific-forgery biases, we have identified a new form of bias, namely \textit{spatial bias}. It refers to the detector's inclination to inertly anticipate observing structural forgery clues at the image center. For example, as shown in the third row of Fig.~\ref{fig:bias}, the model consistently focuses on the image center and ignores the localized and shifted forgery clues.
We believe this issue adversely affects both the effectiveness and robustness of deepfake detection. 
\begin{itemize}
    \item \textbf{Impact on Effectiveness}: inertly focusing on the central area diminishes the sensitivity of the detector to forgery artifacts, which further encourages the model to completely rely on the simple features, while neglecting similarly predictive complex features~\cite{simple-bias}. Moreover, the anticipation of structural artifacts leads to limited performance in detecting local forgeries.
    \item \textbf{Impact on Robustness}: it is evident that variations in preprocessing or camera movement are prevalent in practical applications of deepfake detection. Spatial bias compromises the network robustness against such spatial deviation, as detectors primarily concentrate on the image center, yet spatial deviations displace faces and artifacts away from the focal region. Experiments about robustness could be found in Sec.~\ref{sec:robust}
\end{itemize}

\section{Proposed Method}
\subsection{Overall Data-level Debiasing Framework}
To explicitly remove bias at the data level, we propose a unified framework named ED$^4$ with two essential components. Firstly, we propose ClockMix with arbitrary faces to remove content and specific-forgery biases within the samples. Secondly, we introduce the Adversarial Spatial Consistency Module to address the spatial bias. Our method explicitly extends the training distribution with rich compositions of identity, background, manipulation traces, and spatial arrangements, 
allowing the detector to learn generalizable forgery representations directly and \reG{mitigate the adverse effects of model bias}.
\reG{As the overall architecture in Fig.~\ref{fig:main_arch}}, ED$^4$ is a plug-and-play module that can be directly implemented into the backbone feature extractor. \reT{Namely, ED$^4$ can be applied to the training samples without disrupting the original DFD training pipeline of the backbone network. While during the inference stage, our method requires only the ED$^4$-trained feature extractor (same network architecture as the backbone) to perform detection.}
\subsection{ClockMix} 
\label{sec:ClockMix}


To disrupt the shortcut mapping associated with the specific-forgery, background, and identity, 
we propose ClockMix 
to randomly integrate faces and backgrounds of arbitrary images.
ClockMix performs sector-based mosaicing centered on the face, which is analogous to the rotation of clock hands. Specifically, ClockMix \reG{initially performs} face alignment to \reG{guarantee} that the center of each face is consistently located at the central point of each image according to the face landmarks. Then, we introduce a ``clock hand'' ray $r_h$ starting at the central point $(\delta_x,\delta_y)$ and define the angle between $r_h$ and $r_{base}$ as $\rho$, where $r_{base}$ is a baseline ray starting at the central point toward a random direction. The area in each image swept by $r_h$ upon rotating to $\rho$ will be replaced by the corresponding area from other images within the same mini-batch.
To achieve this effect, we calculate a swept-area matrix $\mathbf{M}$ to record the angles between each $r_{(i,j)}$ and the vertical upward-oriented ray $r_v$, where $r_{(i,j)}$ denotes the ray starting at the central point and passing the pixel located at point $(i,j)$. Hence, each element in $\mathbf{M}$ can be calculated as:
\begin{equation}
\mathbf{M}(i, j) = \left(\frac{180}{\pi} \arctan2\left(\delta_y - i, j - \delta_x\right)\right) \mod 360 ,
\end{equation}
where $\arctan2(y, x)$ denotes the angle in radians between the positive x-axis and the ray to the point $(x, y)$. Considering the angle deviation from $r_{base}$ to $r_v$, the swept-area matrix for randomly generated $r_{base}$ can be written as:
\begin{equation}
\mathbf{M}_{base} = (\mathbf{M}-\rho_{base}) \mod 360,
\end{equation}
where $\rho_{base}$ denotes the deviation angle of $r_{base}$ from $r_v$.
By leveraging $\mathbf{M}_{base}$, we can conduct ClockMix on arbitrary images $\mathbf{I}_a$ and $\mathbf{I}_b$ to obtain the mixed image $\mathbf{I}_{ab}$, which can be written as:
\begin{equation}
\begin{aligned}
\mathbf{I}_{ab} &=ClockMix(\mathbf{I}_a,\mathbf{I}_b,\rho_1) \\
&= \mathbf{I}_{a} \odot (\mathbf{M}_{base} > \rho_1) + \mathbf{I}_{b} \odot (\mathbf{M}_{base} \leq \rho_1),
\end{aligned}
\end{equation}
where $j\odot k$ yields the value $j$ when the condition $k$ is true, and zero otherwise.
ClockMix introduces the simultaneous mixing of faces and backgrounds without damaging the fine-grained textures of images including forgery artifacts. Moreover, by aligning with the center of the face, the overall distribution of a single image still maintains the appearance of a normal face, with the correct number and peripheral relationships of facial attributes.
Then, to enhance the removal of content and specific-forgery biases, we iteratively conduct ClockMix to enable the mixing of multiple faces and backgrounds into one image.
Given $\mathbf{I}_{ab}$ and $\mathbf{I}_{c}$, the mixture of three images is made by
\begin{equation}
    \mathbf{I}_{abc} = ClockMix(\mathbf{I}_{ab},\mathbf{I}_{c},\rho_2).
\end{equation}
To ensure the region from $\mathbf{I}_{b}$ in $\mathbf{I}_{ab}$ is not completely covered by $\mathbf{I}_c$, we always let $\rho_2 < \rho_1$. Analogously, ClockMix can sequentially achieve the mixing of $n$ arbitrary images. 

Regarding the labels of the mixed images, there are three conditions, that is, \textit{Real+Real}, \textit{Real+Fake}, and \textit{Fake+Fake}. 
We posit that any mixed image containing forgery should be considered Fake, otherwise, it is deemed as Real. 
Hence, the output label of the mixed image $\mathbf{I}_{ab}$ is defined by: 
\begin{equation}
\label{eq:clockmixlabel}
    y_{ab} = 1- (1-y_{a})(1-y_{b}),
\end{equation}
where $y_{a} \in \{0, 1\}$ and $y_{b} \in \{0, 1\}$ are the labels of $\mathbf{I}_{a}$ and $\mathbf{I}_{b}$, respectively. In this paper, we use the label $y=0$ for real images and $y=1$ for fake images.

Compared with existing fake synthesizing methods with particular image pair, face region, and label selection \cite{adv,leakage,xray}, or limited identity and forgery participation \cite{SBI}, our ClockMix performs more thorough synthesis with arbitrary images. The deepfake detector is allowed to see mixtures of different identities and backgrounds in both real and fake samples, and the co-occurrence of multiple manipulation clues in one fake image, leading to better removal of the content and specific-forgery bias.


\begin{figure}[t]
    \centering
    \includegraphics[width=0.95\linewidth]{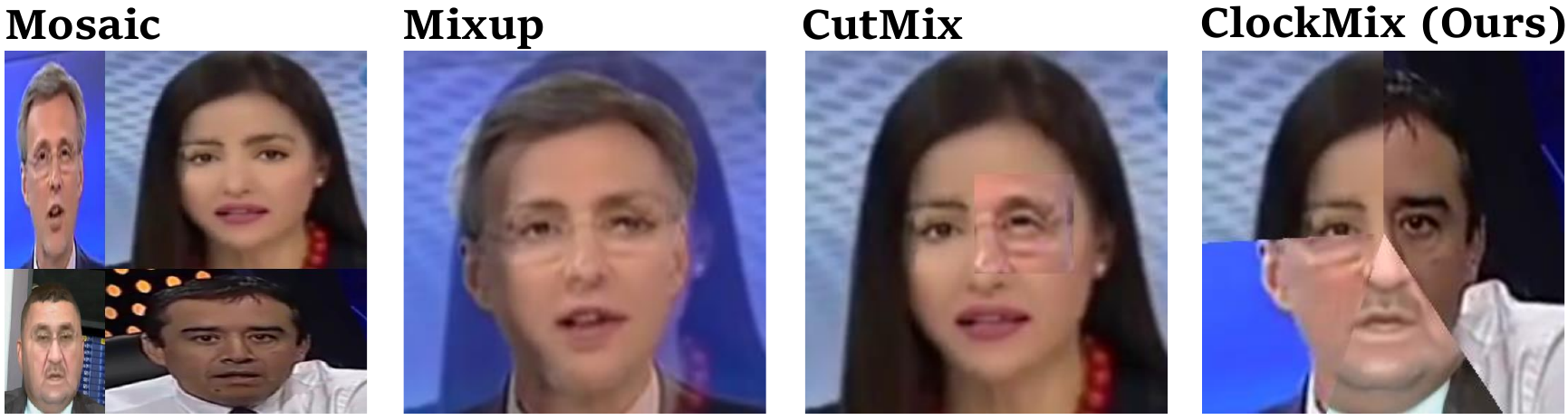}
    \caption{Illustration and comparison of ClockMix and previous popular mixing-based methods.}
    \label{fig:clockmix}
\end{figure}
\textbf{Superiority over existing mixing-based methods.} We \reG{observe} that the proposed ClockMix may share similarities with popular \reG{pixel-wise mixing data augmentation} strategies.
, including Mosaic~\cite{yolov4}, Mixup \cite{mixup}, and Cutmix \cite{cutmix}. As shown in Fig.~\ref{fig:clockmix}, existing strategies are not suitable for debiasing in the deepfake detection task. Specifically, 
the superiority of ClockMix can be demonstrated by answering two questions: i) What kind of mix is more appropriate for deepfake detection? ii) How to assign the labels for mixed images? 
\reG{In the first case,} Mosaic introduces multiple intact faces into one single sample, \reG{thereby disrupting} the high-level statistical distribution of the \reG{training data} and \reG{confusing the detectors}.
Mixup can lead to the overlap of local textures between faces and thus obfuscate the details of forgery artifacts. Cutmix cannot guarantee even mixtures of both face regions and backgrounds from different images simultaneously,  thereby reducing its effectiveness in addressing content bias. By contrast, ClockMix conducts sector dividing based on the face center, gaining better control over selecting useful regions and assigning labels. 
As for the second question, 
Mosaic is designed for object detection, therefore it is not intuitive to assign a classification label.
Mixup and CutMix introduce a mixing label strategy with linear interpolation of 
\begin{equation}
\label{eq:mixuplabel}
   \Tilde{y}_{ab}= \lambda \cdot y_a + (1-\lambda) \cdot y_b,
\end{equation}
with $\lambda \in \left[0,1\right]$.
However, in experiments, we observed that linear interpolation of the one-hot labels actually weakens the detector's ability to identify subtle forgery traces. This is consistent with the basic intuition: an image should be classified as fake even with a small region of manipulated faces. By assigning the hard labels to mixed images, ClockMix can greatly increase the sensitivity of the detector to forgery artifacts. 
\begin{figure}
    \centering
    \includegraphics[width=1\linewidth]{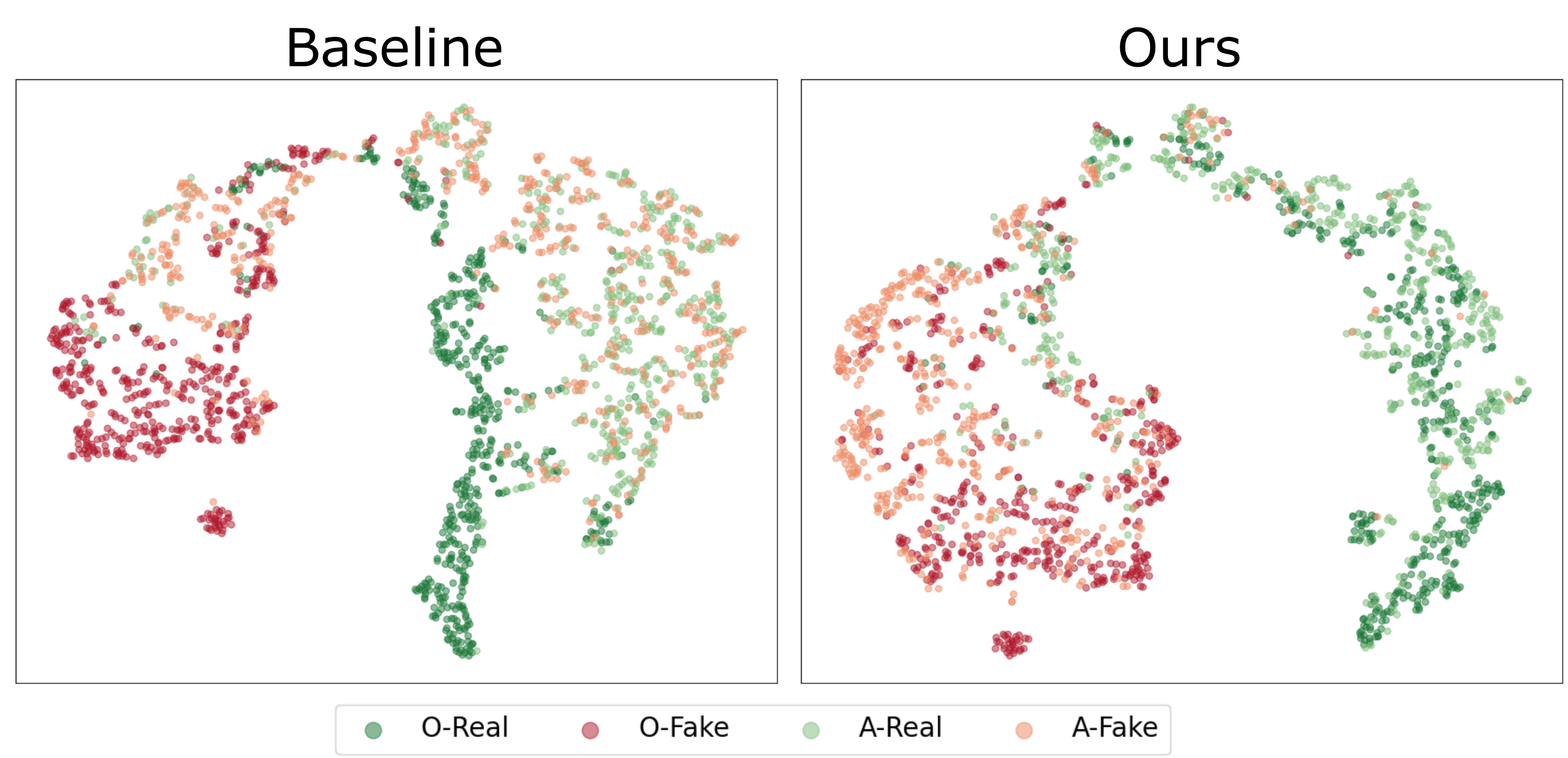}
    \caption{\reT{T-SNE~\cite{t-SNE} visualization for feature distributions. O denotes the Original samples and A denotes the ED$^4$-Augmented samples. It could be observed that Ours is Real/Fake separated and A/O are indispensable, while Baseline's A Real/Fake are distributed next to O-Real. This indicates that the seaming artifacts introduced by our method are ignored by the network, where the distinguished fake and real feature distributions each comprise ClockMix-synthesized and real-world samples of their respective class.}}
    \label{fig:tsne}
\end{figure}

More interestingly, we find that referring to the mixture of real images as \textit{real} also plays a critical role in improving the model performance. As ClockMix inevitably introduces unappealing seams with different image pieces, the deepfake detector might be prone to \reG{relying on the seam} artifacts for determining authenticity. When we perform mixing on real images with the assignment of real labels, \reT{along with introducing a portion of original samples into training, the detector can get rid of the impact of unnatural seams, and meanwhile, distinguish real from fake with improved accuracy. In Fig.~\ref{fig:tsne}, we conduct experiments on both original samples (\textit{i.e.}, O-Real and O-Fake) and augmented samples (\textit{i.e.}, A-Real and A-Fake with seaming artifacts), and visualize the feature distributions of baseline and our method. The results show that our detector could distinguish both O-Real/Fake and A-Real/Fake without splitting each other, demonstrating that the seaming artifacts are ignored by our feature extractor.}

\reT{In summary, the advantages of ClockMix among existing mixing-based augmentations are twofold:
\begin{itemize}
  \item \textbf{Regionally Controlled Sector-Based Mixing.} ClockMix partitions each image into angular sectors centered on the facial midpoint, enabling targeted extraction and recombination of both facial and background regions. This approach preserves critical forgery artifacts while avoiding the high-level distribution distortion introduced by Mosaic, the texture blurring of Mixup, and the unbalanced region sampling of CutMix.
  \item \textbf{Hard-Label Assignment with Real-Data Regularization.} By assigning binary ``real'' or ``fake'' labels to mixed samples -- rather than employing linear interpolation of one-hot vectors -- ClockMix maintains maximal sensitivity to subtle manipulation traces. Furthermore, labeling real–real mixtures as genuine regularizes the detector against seam-based artifacts, thereby enhancing robustness and overall classification accuracy. Furthermore, this labeling strategy allows conducting ClockMix to mix multiple images instead of two images, which enhances the diversity of mixing results.
\end{itemize}}

Please refer to the \textit{Ablation Study} for more experimental results and illustrations.


\subsection{Adversarial Spatial Consistency Module}
\begin{figure}[tbp]
    \centering
    \includegraphics[width=\linewidth]{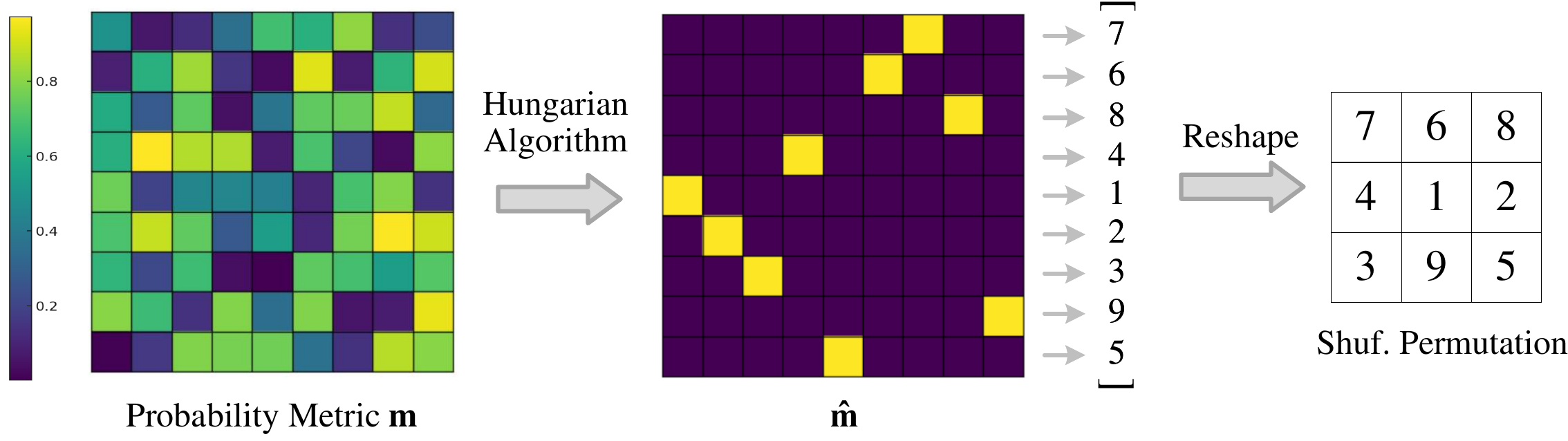}
    \caption{Visualized example of transforming generator output to shuffle permutation via linear programming.}
    \label{fig:vis}
\end{figure}

As shown in Fig.~\ref{fig:main_arch}, we propose the Adversarial Spatial Consistency Module (AdvSCM) to address spatial bias in the extracted feature.
The adversarial consistency strategy refers to the generator (enlarging spatial inconsistency) \textit{vs.} the backbone extractor (reducing feature inconsistency), which closely resembles the generator \textit{vs.} discriminator in the generative adversarial network.
By achieving this, the backbone can learn to extract features that are \textit{spatial-agnostic}, thus avoiding spatial bias. In AdvSCM, we first introduce spatial shuffle~\cite{shuffle} as the base operation, which does not undermine the texture and fine-grained information but only produces spatial inconsistency. 
Then, given an input image $\mathbf{I}$, two shuffled versions of $\mathbf{I}$ are generated by the random generator $\mathcal{G}_r(\cdot)$ and the adversarial generator $\mathcal{G}_a(\cdot;\theta_a)$, respectively, where $\theta_a$ is the learnable parameter in $\mathcal{G}$.
Without learnable parameters, $\mathcal{G}_r$ relocate blocks and reassemble them to obtain the shuffled image $\mathbf{I}_{s1}=\mathcal{G}_r(\mathbf{I})$ with \textit{random} granularity and permutation. In contrast, $\mathcal{G}_a$ is a neural network that inputs the concatenation of $\mathbf{I}$ and $\mathbf{I}_{s1}$, and outputs a probability metric $\mathbf{m}=\mathcal{G}_a((\mathbf{I},\mathbf{I}_{s1}); \theta_a) \in \mathbb{R}^{N \times N}$. Namely, given a granularity $g$, the number of divided patches is $N=g^2$. \reT{To be more specific, $g$ represents the number of divisions along both the horizontal and vertical axes of an image, such that the image is then divided into $N=g^2$ patches.} Hence, the shape of the output possibility metric $\mathbf{m}$ is $N\times N$, where the first $N$ refers to the number of patches in the shuffled image, the second one refers to the possible shuffled location of patches. Such an effect can be easily achieved by deploying a linear layer that outputs $N \times N$ after the network backbone and then reshaping the output. \reT{In implementation, the feature extraction backbone of $G_a$ remains shared across all different $N$. Only the number of channels in the final linear layer is adjusted to accommodate the current $N$ value.}
However, directly using $\mathbf{m}$ to guide the shuffling of images could be problematic. Specifically, each region should be assigned to a unique patch, that is, the predicted region index for each patch must be distinct. In fact, this mutual exclusivity of indices aligns with solving polynomials, except we are maximizing polynomials instead of minimizing them, which could be easily transformed by adding a minus sign to $\mathbf{m}$. Therefore, the problem can be formulated as:
\begin{equation}
\text{minimize } \sum_{i=1}^{N} \sum_{j=1}^{N} - m_{i,j} \hat{m}_{i,j}, 
\end{equation}
where $x_{i,j} \in \{0, 1\}$ and
\begin{equation}
    \prod_{i=1}^{N}\sum_{j=1}^{N}\hat{m}_{i,j}=1, \space \prod_{i=1}^{N}\sum_{i=1}^{N}\hat{m}_{i,j}=1.
\end{equation}
Subsequently, we utilize the Hungarian algorithm to solve this polynomial and hence convert $\mathbf{m}$ into $\mathbf{\hat{m}}$. 
Then, the max indices in each row of $\mathbf{\hat{m}}$ could be directly used as the permutation that can guide the shuffling.
We provide a visualized example for a better illustration of the aforementioned process in Fig~\ref{fig:vis}.

Both $\mathbf{I}_{s1}$ and $\mathbf{I}_{s2}$ are then input into the backbone extractor \( \mathcal{E}(\cdot; \theta_e) \) to obtain their respective features $\mathbf{F}_{s1}, \mathbf{F}_{s2}$. 
The optimization target of $\theta_e$ is minimizing the distance between $\mathbf{F}_{s1}$ and $\mathbf{F}_{s2}$, which can be formulated as:
\begin{equation}
    \theta_{e}'=arg\underset{\theta_{e}}{\text{min}} D, \label{argmin}
\end{equation}
where $D=L_1(\mathbf{F}_{s1},\mathbf{F}_{s2})$ represents L1 distance between features, $\theta_{e}'$ is the updated parameter, and $\theta_a$ should be frozen. Such optimization can be directly implemented via backpropagation. Similarly, $\theta_a$ should be optimized for maximizing feature distance:
\begin{equation}
    \theta_{a}'=arg\underset{\theta_{a}}{\text{max}} D,
\end{equation}
where $\theta_e$ should be frozen. Notably, the gradient flow for optimizing $\theta_a$ is broken by non-differentiable operations like sampling and the Hungarian algorithm. Therefore, we approximate the gradient calculation via \reG{the} Reinforce algorithm~\cite{reinforce}:
\begin{equation}
    \theta_{a}'=\theta_{a}+\frac{\epsilon}{K}\sum_{k=1}^{K}D\nabla_{\theta_a}\text{log}(p), \label{argmax2}
\end{equation}
where $K$ denotes the size of mini-batch, $\epsilon$ is the learning rate, and 
\begin{equation}
 p=\frac{1}{N}\sum_{i=1}^N\sum_{j=1}^N m_{i,j} \hat{m}_{i,j},
\end{equation}
where $m_{i,j}$ and $\hat{m}_{i,j}$ refer to the entry located in the $i$-th row and $j$-th column of $\mathbf{m}$ and $\mathbf{\hat{m}}$. Both $m_{i,j}$ and $\hat{m}_{i,j}$ can be interpreted as the probability that the patch at the $i$-th position is shuffled to the $j$-th position. 

By introducing AdvSCM, the adversarial generator learns to predict \reG{a} shuffle configuration that maximizes the spatial inconsistency between generated shuffled images, thus allowing the backbone extractor to more robustly learn spatial-agnostic features via consistency constraints. \reT{The superiority of the proposed AdvSCM could be summarised as follows:
\begin{itemize}
  \item \textbf{First Explicit Spatial-Bias Mitigation via Shuffle Consistency and Adversarial Adaptation.} AdvSCM is the first method explicitly designed to counteract the detector’s central-region bias: it employs a shuffle-consistency strategy to disrupt over-reliance on structural forgery cues at the facial center and integrates an adversarial module to compel robust adaptation to manipulations at arbitrary spatial locations.
  \item \textbf{Multi-Granularity Feature Consistency Across Rearrangements.} Unlike prior spatial-aware approaches that rely on implicit spatial transformations, relational constraints, or frame thumbnail rearrangements, AdvSCM enforces consistency at multiple feature scales between two distinct spatial permutations of the same image, thereby more effectively disentangling genuine content features from positional bias.
\end{itemize}}

\begin{table*}[ht]
\centering

\caption{Cross-dataset evaluations (AUC) from FF++ to CDFv1, CDFv2, DFD and DFDC based on DeepfakeBench~\cite{deepfakebench} (Protocol 1). C-Avg. denotes the average value of cross-dataset results. The best results are highlighted in \textbf{bold}.} \label{res-DFB}
\centering
\small
\begin{tabular}{l|c|ccccccc} \toprule
Method  &Venue    &   FF++~\cite{FF++}     & CDFv1~\cite{Celeb-df}& CDFv2~\cite{Celeb-df} & DFD~\cite{dfd}  & DFDC~\cite{dfdc}  & C-Avg. \\ \midrule
Xception \cite{xception} &CVPR'17&   0.9637  &  0.7794     &    0.7365   &   0.8163  &   0.7077   &  0.7600 \\
FWA \cite{FWA}     & CVPRW'18&  0.8765   & 0.7897 & 0.6680 & 0.7403 & 0.6132&            0.7028\\
Meso4 \cite{mesonet}     & WIFS'18  &0.6077 &      0.7358&       0.6091&       0.5481&     0.5560&      0.6123\\
Capsule \cite{capsule}  &ICASSP'19 &  0.8421 &      0.7909&       0.7472&       0.6841&     0.6465&        0.7172\\
EfficientNetB4 \cite{effnet} & ICML'19  & 0.9567 &      0.7909&       0.7487&       0.8148&     0.6955&       0.7625\\
CNN-Aug \cite{cnn-Aug}    & CVPR'20 & 0.8493 &      0.7420&       0.7027&       0.6464&     0.6361&     0.6818\\
X-ray \cite{xray}   &CVPR'20&  0.9592  &   0.7093 & 0.6786 & 0.7655 & 0.6326&            0.6965\\
FFD \cite{ffd}   & CVPR'20&  0.9624 &   0.7840 & 0.7435 & 0.8024 & 0.7029&           0.7582\\
F3Net \cite{F3Net}   &ECCV'20&  0.9635 &   0.7769 & 0.7352 & 0.7975 & 0.7021&         0.7530\\ 
SPSL \cite{spsl}    &CVPR'21&  0.9610 &  0.8150 & 0.7650 & 0.8122 & 0.7040 &         0.7741\\ 
SRM \cite{srm}     &CVPR'21&   0.9576  &  0.7926 & 0.7552 & 0.8120 & 0.6995 &         0.7648\\  
CORE \cite{core}    &CVPRW'22& 0.9638  &  0.7798 & 0.7428 & 0.8018 & 0.7049&         0.7573\\ 
Recce \cite{recce}  & CVPR'22& 0.9621 &  0.7677 & 0.7319 & 0.8119 & 0.7133&         0.7562\\ 
SLADD \cite{adv}  &CVPR'22 &   0.9691      & 0.8015 & 0.7403 & 0.8089 & 0.7170 &         0.7669\\
SBI \cite{SBI}   &CVPR'22&   0.9747   & 0.8371 & 0.7911 & 0.8112 & 0.7275 &         0.7917\\  
UIA-VIT \cite{uia} &ECCV'22&0.9655 &0.7925&0.8002&0.8279&0.7038&0.7811 \\
IID \cite{huang2023implicit}&CVPR'23&0.9743&0.7578&0.7687&0.7935&0.6951& 0.7538 \\
UCF \cite{ucf}   &  ICCV'23  &  0.9705 &  0.7793 & 0.7527 & 0.8074 & 0.7191&  0.7646\\ 
\midrule
Liang \etal \cite{Disentangle}&ECCV'22& - &  0.706   & - & 0.829 & 0.700 &         -\\ 
LSDA \cite{lsda}     & CVPR'24 &  - &  0.867 & 0.830 & \textbf{0.880} & 0.736&  0.828\\ 
\midrule
ED$^4$ (Ours)       &   -   &  \textbf{0.9806}     &  \textbf{0.8871} & \textbf{0.8394} & 0.8765 & \textbf{0.7427} &         \textbf{0.8364}\\ 
\bottomrule
\end{tabular}
\end{table*}

\begin{table*}[htbp]
\centering
\caption{Cross-dataset evaluations on more various and advanced datasets (Protocol 2). The metric is frame-level AUC and all models are trained on FF++ (HQ)~\cite{FF++}.} \label{tab:auc}
\small
\begin{tabular}{lcccccccccc}\toprule
Methods  & WDF & FAVC & DiffSwap & E4S & BlendFace & UniFace & DFL  & \reT{SDv21} & Avg.  & \reT{\#Top1}\\ \midrule
EfficientNet~\cite{effnet} & 0.7275 & 0.8404 & 0.7956 & 0.6514 & 0.7813 & 0.7775 & 0.7347 & \reT{0.5581} & \reT{0.7333}& \reT{0} \\
Xception~\cite{xception} & 0.7705 & 0.8681 & 0.8110 & 0.6554 & 0.7911 & 0.8282 & 0.7199 & \reT{0.6145} & \reT{0.7573}& \reT{0} \\
SPSL~\cite{spsl} & 0.7023 & 0.7689 & 0.7784 & 0.5987 & 0.6936 & 0.6915 & 0.6548 & \reT{0.5877} & \reT{ 0.6845}& \reT{0} \\
UCF~\cite{ucf} & \textbf{0.7738}& 0.8631 & 0.8511 & 0.6916 & 0.7850 & 0.7867 & 0.7433 & \reT{0.5910} & \reT{0.7607}& \reT{1}\\
\reT{DCL~\cite{dcl}} & \reT{0.7401} & \reT{0.6169} & \reT{0.8245} & \reT{\textbf{0.7611}} & \reT{0.7992} & \reT{0.7145} & \reT{0.7394} & \reT{0.6315} & \reT{ 0.7284}& \reT{1} \\
\reT{TALL~\cite{tall}} & \reT{0.7428} & \reT{0.8031} & \reT{0.7971} & \reT{0.7030} & \reT{0.8320} & \reT{0.8310} & \reT{0.6761} & \reT{0.6501} & \reT{0.7544}& \reT{0} \\
\reT{CDFA~\cite{cdfa}} & \reT{0.7437}& \reT{0.8861} & \reT{0.5479} & \reT{0.6256}& \reT{0.6840}& \reT{0.7576}& \reT{0.6139} & \reT{0.6070} & \reT{0.6832}& \reT{0}\\
\reT{RAM~\cite{ram}} & \reT{0.6991} & \reT{0.7742} & \reT{0.5991} & \reT{0.5991} & \reT{\textbf{0.8679}} & \reT{\textbf{0.8323}} & \reT{0.7530} & \reT{0.5071} & \reT{0.7084} & \reT{2} \\ \midrule
ED$^4$ (Ours) & 0.7682 & \textbf{0.8879} & \textbf{0.8781} & 0.7201 & 0.8511 & 0.8280 & \textbf{0.7794} & \reT{\textbf{0.6704}} & \reT{\textbf{0.7979}}& \reT{5} \\ \bottomrule
\end{tabular}

\end{table*}

\subsection{Detection Loss}
To classify forgeries, the input image $\mathbf{I}$ is processed by the detector and obtains a detection result $y'$. 
The ground-truth label $y$ is calculated following Eq.~\ref{eq:clockmixlabel} if $\mathbf{I}$ is mixed. 
Since ED$^4$ requires no modification on the backbone network, the forgery detection loss $L_d$ can be simply measured with the binary cross-entropy loss as
\begin{equation}
    L_d(y', y) = -[y \log(y') + (1 - y) \log(1 - y')],
\end{equation}
which enables the network to identify the forgery images.

\section{Experimental Results}
\subsection{Experimental Setting}
\subsubsection{Datasets} Given that the generalization issue is the major challenge for research, we apply an \reG{abundance} of advanced and widely used deepfake datasets in our experiments.
FaceForensics++ (FF++)~\cite{FF++} is constructed by four forgery methods including Deepfakes (DF)~\cite{df}, Face2Face (F2F)~\cite{f2f}, FaceSwap (FS)~\cite{fs}, and NeuralTextures (NT)~\cite{nt}. Meanwhile, it includes three different compression quality levels, that is, RAW, High-Quality (HQ), and Low-Quality (LQ). We employ FF++ (HQ) as the training dataset for all experiments in our paper. For cross-dataset evaluations, we introduce both classical and advanced datasets, including Celeb-DF-v1 (CDFv1)~\cite{Celeb-df}, Celeb-DF-v2 (CDFv2)~\cite{Celeb-df}, DeepFakeDetection (DFD)~\cite{dfd}, DeepFake Detection Challenge (DFDC)~\cite{dfdc}, DF40~\cite{df40}, DiffusionFace~\cite{diffusionface}, and more. These datasets have covered all types of forgery categories, including Face-Swapping, Face-Reenactment, Face-Editing, and Entire Face Synthesis~\cite{df40}.
\\
\\
\subsubsection{Implementation Details} 
For preprocessing and training, we strictly follow the official code and settings provided by DeepFakeBench~\cite{deepfakebench} to ensure fair comparison. DeepFakeBench is a comprehensive benchmark tailored for deepfake detection, which proposes a unified framework for pre-processing, training, and testing. They also reproduced the SoTA methods with consistent training hyper-parameters. To be specific, Dlib~\cite{dlib} is used for face extraction and alignment with a cropping margin of 15\%. 32 frames are extracted from each video and all frames are cropped to the size of 256 $\times$ 256.  For training, traditional data augmentations are deployed, including rotating, Gaussian noise, saturation adjusting, and quality adjusting. 
For ClockMix, the number of mixing images is randomly selected from $\{1,2,3,4\}$, and $\rho$ is randomly selected from 45 to 315. Our method takes Xception~\cite{xception} as the backbone for both adversarial generator and extractor and they are initialized by the parameters pre-trained on ImageNet \cite{imagenet}. The Adam optimizer is used with a learning rate of 0.0002, \reG{an epoch of 10, an input size of 256 $\times$ 256, and a batch size of 32.}
\textit{Frame-level} Area Under Curve (AUC) and Equal Error Rate (EER)~\cite{deepfakebench} are applied as the evaluation metrics of experimental results. All experiments are conducted on one NVIDIA Tesla V100 GPU.

\subsection{Cross-dataset Evaluation}
\subsubsection{Based on DeepFakeBench (Protocol 1)} In Tab.~\ref{res-DFB}, we provide extensive comparison results with existing state-of-the-art (SOTA) deepfake detectors based on DeepFakeBench~\cite{deepfakebench}. Specifically, all methods are trained on FF++ (HQ) and tested on other datasets. The methods in the upper part of the table are within the benchmark with the exact same experimental setting as our method. Hence, we directly \textit{copy} their results from DeepfakeBench. Notably, more SOTA methods are related to our method but are not included in the DeepFakeBench. Therefore, the results of Liang \etal~\cite{Disentangle} are reproduced by \cite{ucf}. LSDA~\cite{lsda} is the new SOTA based on DeepFakeBench, hence we directly copy the results from their paper.
It can be observed that our method consistently outperforms the earlier deepfake detectors based on data synthesis techniques (\textit{i.e.}, FWA, X-ray, SLADD, and SBI) across all evaluated datasets.  Meanwhile, compared with the implicit disentanglement methods (\textit{i.e.}, Liang \etal and UCF), our method also exhibits superior effectiveness. These results demonstrate the superiority of the proposed explicit data-level debiasing approach over previous SOTA methods.

\subsubsection{More Advanced Datasets (Protocol 2)} 
Here, we provide results of our method on \textbf{seven} more datasets, including WildDeepfake~\cite{wilddeepfake} (WDF), FakeAVCeleb~\cite{fakeavceleb} (FAVC), \{E4S~\cite{e4s}, BlendFace~\cite{blendface}, UniFace~\cite{uniface}, DeepfaceLab~\cite{deepfacelab} (DFL)\} from DF40~\cite{df40}, and \{DiffSwap~\cite{diffswap}, \reT{SDv21~\cite{sd}}\} from DiffusionFace~\cite{diffusionface}. Considering most baseline methods from DeepfakeBench~\cite{deepfakebench} have not been reproduced on these advanced datasets, we carefully reproduce EfficientNet~\cite{effnet}, Xception~\cite{xception}, SPSL~\cite{spsl}, UCF~\cite{ucf}, \reT{DCL~\cite{dcl}, TALL~\cite{tall}, CDFA~\cite{cdfa}, and RAM~\cite{ram}}, and then evaluate them on these datasets for comparison. As shown in Tab.~\ref{tab:auc}, it can be observed that our methods can be generalized to various datasets with superior effectiveness. 
\subsection{Ablation Study}
\begin{table}[t]
\centering
\caption{The impact of ClockMix (CM) and Adversarial Spatial Consistency Module (AdvSCM).} \label{tab:abl1}
\footnotesize
\centering
\begin{tabular}{l@{\hspace{0.1cm}}c@{\hspace{0.2cm}}c@{\hspace{0.2cm}}c@{\hspace{0.2cm}}c@{\hspace{0.2cm}}c@{\hspace{0.2cm}}c}\toprule
\multirow{2}{*}{Method} & \multicolumn{2}{c}{\hspace{-0.15cm} DFD}& \multicolumn{2}{c}{\hspace{-0.15cm} CDFv2} & \multicolumn{2}{c}{\hspace{-0.15cm} DFDC} \\ \cmidrule(lr){2-3} \cmidrule(lr){4-5} \cmidrule(lr){6-7}
                     & AUC         & EER   & AUC         & EER         & AUC         & EER        \\ \midrule
Baseline         &0.7994&0.2738&     0.7454    & 0.3209 &   0.7089    &   0.3621     \\
w/o CM         &0.8336&0.2280&     0.7802    & 0.2907 &   0.7105    &   0.3517     \\
w/o AdvSCM              &0.8200&0.2499&       0.7712       & 0.3143 &   0.7296   &  0.3327    \\
Ours                & \textbf{ 0.8765 } &\textbf{ 0.1939} & \textbf{0.8394}& \textbf{0.2283}  &  \textbf{0.7427} &   \textbf{0.3048} \\ \bottomrule
\end{tabular}
\end{table}
\subsubsection{Impact of Different Proposed Components}
In Tab.~\ref{tab:abl1}, we assess the effectiveness of each proposed component. The detector trained without all proposed components is denoted by Baseline. The results on all evaluated datasets and metrics demonstrate that each component essentially contributes to the effectiveness of our method. 

\begin{table}[t]
\centering
\caption{Comparison of different mixing and label-assignment strategies. HL indicates assigning hard labels with Eq.~\eqref{eq:clockmixlabel}, and SL denotes calculating soft labels with Eq.~\eqref{eq:mixuplabel}. MRF denotes assigning the mixture of real images with the \textit{fake} label.} \label{tab:abl-CM}
\centering
\footnotesize
\begin{tabular}{l@{\hspace{0.14cm}}c@{\hspace{0.14cm}}c@{\hspace{0.28cm}}c@{\hspace{0.14cm}}c@{\hspace{0.28cm}}c@{\hspace{0.14cm}}c}\toprule
\multirow{2}{*}{Method} & \multicolumn{2}{c}{\hspace{-0.15cm} DFD}& \multicolumn{2}{c}{\hspace{-0.15cm} CDFv2} & \multicolumn{2}{c}{\hspace{-0.15cm} DFDC}\\ \cmidrule(lr){2-3} \cmidrule(lr){4-5} \cmidrule(lr){6-7}
                 &   AUC    & EER    & AUC         & EER         & AUC         & EER        \\ \midrule
                 CM-MRF  &  0.6061  & 0.4571 & 0.5419& 0.4833  &  0.5351 &   0.4877         \\  \midrule
Mosaic-HL        &  0.7101  & 0.3479 & 0.5939      & 0.4638      &  0.5714     &   0.4312   \\
Mosaic-SL        &  0.7915  & 0.2678 & 0.7549      & 0.3101      &  0.6792     &   0.3795    \\
CutMix-HL        &  0.8399  & 0.2124 & 0.7862      & 0.2868      &  0.7231     &   0.3471   \\
CutMix-SL        &  0.7534  & 0.2910 & 0.7357      & 0.3279      &  0.6322     &   0.4301    \\
Mixup-HL         &  0.8167  & 0.2518 & 0.7575      & 0.3023      &  0.6959     &  0.3601    \\
Mixup-SL         &  0.8013  & 0.2802 & 0.7464      & 0.3173      &  0.7009     &  0.3595    \\  \midrule

CM-SL  &  0.8104  & 0.2593 & 0.7953& 0.2799  &  0.7191 &   0.3584         \\ 
CM-HL(Ours)              & \textbf{ 0.8765 } &\textbf{ 0.1939} & \textbf{0.8394}& \textbf{0.2283}  &  \textbf{0.7427} &   \textbf{0.3048} \\ \bottomrule
\end{tabular}

\end{table}

\subsubsection{Configurations of ClockMix} \label{Exp:clockmix}
To demonstrate the superiority of the proposed ClockMix, we replace ClockMix with traditional data mixing augmentations, that is, Mosaic~\cite{yolov4}, Mixup~\cite{mixup}\reG{, and CutMix~\cite{cutmix}}. We also investigate the influences of label-assigning strategies, including Hard Label (HL), Soft Label (SL), and assigning the Mixture of Real images with the Fake label (MRF).
In Tab.~\ref{tab:abl-CM}, we first observe that MRF fundamentally destroys the detector's ability to justify forgery clues. This can be attributed to all images with ClockMix being assigned with Fake labels, hence the detector mistakenly takes the ClockMix splicing patterns as the indicator of forgery.
The proposed ClockMix-HL significantly outperforms all ablation variants in performance. 
Image with Mosaic cannot be applied with HL considering the intact information is preserved for each mixed face. Meanwhile, it can be observed that SL severely undermines the effectiveness of CutMix and ClockMix by categorizing the Real+Fake images under an uncertain label, which reduces the confidence in determining fake images with minor forgery artifacts. Since applying Mixup renders the details of forgery artifacts, both HL and SL versions of Mixup compromise the effectiveness of the deepfake detector.
\begin{table}[t]
\centering
\caption{Effectiveness of the proposed adversarial strategy compared with random strategies.} \label{tab:abl-AdvSCM}
\centering
\footnotesize
\begin{tabular}{lc@{\hspace{0.1cm}}c@{\hspace{0.2cm}}c@{\hspace{0.1cm}}c@{\hspace{0.2cm}}c@{\hspace{0.1cm}}c}\toprule
\multirow{2}{*}{Method} & \multicolumn{2}{c}{\hspace{-0.15cm} DFD}& \multicolumn{2}{c}{\hspace{-0.15cm} CDFv2} & \multicolumn{2}{c}{\hspace{-0.15cm} DFDC} \\  \cmidrule(lr){2-3} \cmidrule(lr){4-5} \cmidrule(lr){6-7}
                          & AUC  & EER    & AUC         & EER         & AUC         & EER        \\ \midrule
w/o AdvSCM              &0.8200&0.2499&       0.7712       & 0.3143 &   0.7296   &  0.3327    \\ \midrule
RS                     & 0.8279& 0.2367 & 0.7839      & 0.2917 &     0.7304 &  0.3209    \\
RS+C    &\textbf{0.8795}& \textbf{0.1892}& 0.8033      &  0.2709     &        0.7212     &       0.3434     \\ 
AdvSCM (Ours)                & 0.8765  &0.1939 & \textbf{0.8394}& \textbf{0.2283}  &  \textbf{0.7427} &   \textbf{0.3048} \\ \bottomrule
\end{tabular}
\end{table}
\begin{table*}[htbp]
\centering
\caption{Comparison with more augmentation methods for spatial bias. Local represents the introduction of local forgery. Region denotes introducing region variations. Info-I denotes Informative pixel Integrity.} \label{tab:aug}
\centering
\begin{tabular}{l@{\hspace{0.2cm}}|@{\hspace{0.2cm}}c@{\hspace{0.2cm}}c@{\hspace{0.2cm}}c@{\hspace{0.2cm}}|@{\hspace{0.2cm}}c@{\hspace{0.4cm}}c@{\hspace{0.4cm}}c@{\hspace{0.4cm}}c}\toprule
Method &Local&Region&Info-I& DFD& CDFv2 &DFDC & Avg.\\ \midrule
w/o AdvSCM (Baseline) && &\checkmark&0.8200&0.7712& 0.7296 &0.7736 \\  \midrule
GridMask         &          &\checkmark&          &0.8408 & 0.7838          &   0.7312  & 0.7853   \\
AttrMask         &          &\checkmark&          &0.8005 & 0.7139         &   0.6759  &  0.7301 \\
RandomCrop       &\checkmark&          &         &0.8415 & 0.8150         &   0.7235  &  0.7933 \\  \midrule
Shuffle &\checkmark&\checkmark&\checkmark&  \textbf{0.8765} & \textbf{0.8394}&  \textbf{0.7427} & \textbf{0.8195}\\ \bottomrule
\end{tabular}
\end{table*}
\begin{figure*}[htbp]
    \centering
    \begin{subfigure}[b]{0.32\textwidth}
        \includegraphics[width=\textwidth]{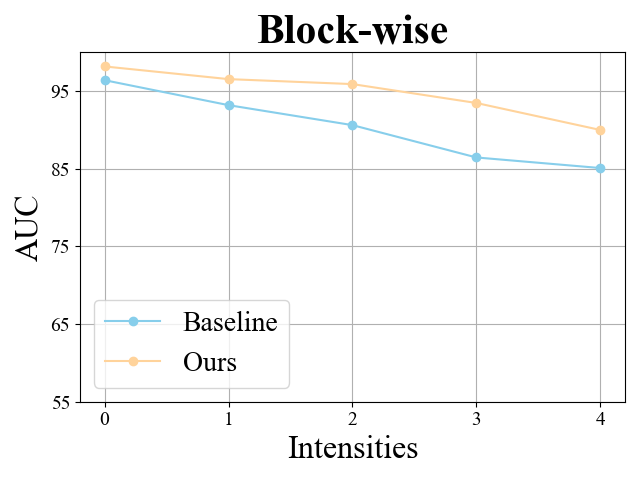} 
    \end{subfigure}
    \hfill
    \begin{subfigure}[b]{0.32\textwidth}
        \includegraphics[width=\textwidth]{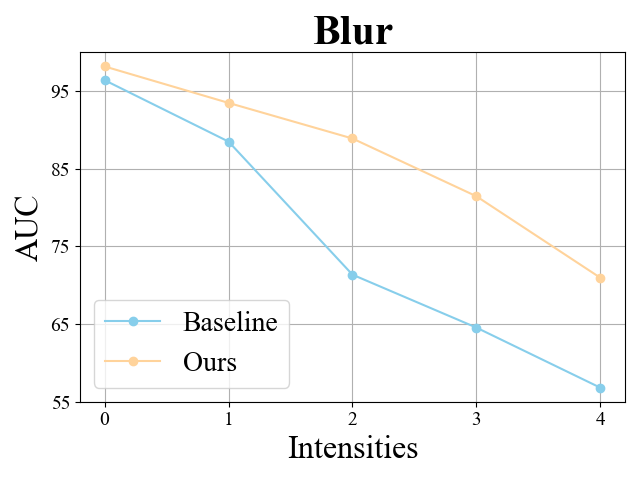} 
    \end{subfigure}
        \hfill
    \begin{subfigure}[b]{0.32\textwidth}
        \includegraphics[width=\textwidth]{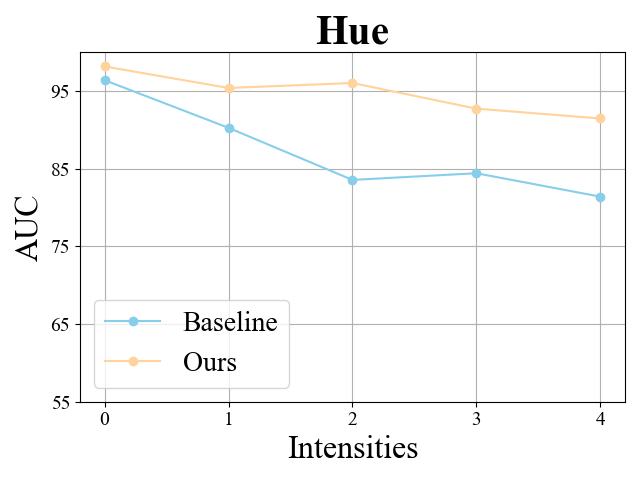} 
    \end{subfigure}
    \caption{Robustness against unseen perturbations with multiple intensities. Block-wise, Blur, and Hue denote block-wise masking, median blur, and hue adjustment, respectively. \reT{The curves across multiple intensities consistently demonstrate that ED4 maintains higher performance than the baseline when confronted with unseen perturbations, thereby highlighting its improved practical applicability for deepfake detection.}} \label{fig:robust}
\end{figure*}
\begin{figure*}[htbp]
    \centering
    \includegraphics[width=1\linewidth]{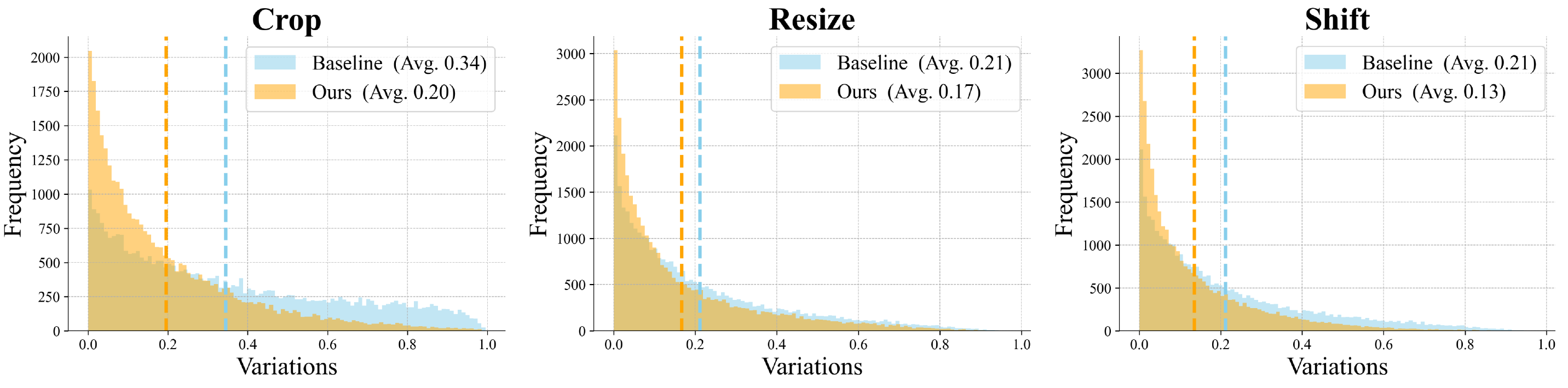}
    \caption{Robustness against spatial deviations. ``Variations" indicates the variation of prediction scores when applying different spatial transformations. The dotted lines represent the average variation. \reT{The results indicate the spatial bias issue in the previous baseline, and also demonstrate the effectiveness of ED$^4$ in spatial bias removal.}}
    \label{fig:robust-spa}
\end{figure*}
\subsubsection{Effect of Adversarial Spatial Consistency Module}
To validate the effectiveness of the proposed adversarial consistency strategy, 
we design two ablation variants: 1) Random Shuffle (RS), where we only deploy random shuffle to the image, and take the shuffled image as the training input. It could be treated as naively deploying Grid Shuffle as a data augmentation operation. 2) Random Shuffle Consistency (RS+C), where we generate two random shuffle versions of an image, and apply Eq.~\eqref{argmin} to encourage their consistency without using Eq.~\eqref{argmax2} for the adversarial strategy. w/o AdvSCM could be treated as the baseline for this ablation.
Tab.~\ref{tab:abl-AdvSCM} shows that only applying RS contributes \reG{trivially} to performance enhancement. RS+C could also improve the generalization performance of \reG{the} detector, which further demonstrates that the shuffle consistency constraint could reduce the spatial bias in the extracted features. However, despite RS+C exhibiting slightly superior effectiveness on the DFD dataset, it cannot stably perform best in different cross-dataset settings. In contrast, deploying AdvSCM enhances the overall cross-dataset generalization ability of deepfake detection.

\subsubsection{Comparison with More Augmentation Methods for Spatial Bias}
As shown in Tab.~\ref{tab:aug}, we introduce three approaches to replace Shuffle that may contribute to spatial bias removal: 1) GridMask~\cite{gridmask}, which is a widely-used augmentation that randomly masks a certain ratio of images. 2) AttrMask, which can mask out certain facial attributes guided by the facial landmarks. 3) RandomCrop, which alters the relative location of faces in the images. The w/o AdvSCM can be treated as the baseline for which no spatial transformation is deployed.
The masking-based methods can break the structural forgery into local ones, but cannot improve the model sensitivity to artifacts at arbitrary regions. While RandomCrop can alter the forgery location to some extent, but still \reG{maintains} the forgery structure. Meanwhile, since masking and cropping lead to the reduction of informative pixels, the performances achieved by these methods are not as pronounced as our S-Shuffle. Especially for AttrMask, despite masking certain facial attributes that may enhance the model sensitivity to local forgery, it significantly reduces the pixels that likely contain forgery information and thus leads to performance degradation. Moreover, this information reduction further leads to inherent inconsistency in extracted features, thereby rendering the employment of Consistency Loss inappropriate. In contrast, deploying Shuffle can uniformly address spatial bias while maintaining informative pixel integrity. Therefore, we use Shuffle as the basic spatial operation to construct the proposed adversarial spatial consistency module.
\begin{table}[tbp]
\centering
\caption{Investigation for mixing images with different labels. Avg. denotes the Average performance in the cross-dataset evaluation.} \label{tab:clock}
\centering
\begin{tabular}{lcccc}\toprule
Method & DFD& CDFv2 &DFDC& Avg.\\ \midrule
Baseline   &0.7994&     0.7454   &   0.7089 &0.7512  \\  \midrule
w/o Real+Real                   &0.7177 & 0.6959          &   0.6361   & 0.6832  \\
w/o Real+Fake                     &0.8481 & 0.8045         &   0.7255  & 0.7927  \\
w/o Fake+Fake   &0.8560&  0.7931    &  0.7247  & 0.7913 \\  \midrule
Ours    &  \textbf{0.8765} & \textbf{0.8394}&  \textbf{0.7427} & \textbf{0.8195}\\ \bottomrule
\end{tabular}
\end{table}

\subsection{The Effectiveness of Conducting ClockMix on Arbitrary Images}
To investigate the influence of conducting ClockMix on arbitrary images, we perform an ablation study with three ablation variants, that is, w/o Real+Real. w/o Real+Fake, and w/o Fake+Fake. As shown in Tab.~\ref{tab:clock}, w/o Real+Real will undermine the performance by indulging the detector to take the seaming patterns as fake. Such a result is similar to the ``mixture of real images with the fake label (MRF)'' that we present in the main paper except the degradation is slighter. Overall, the lack of any label combination in ClockMix can lead to considerable performance degradation, which indicates the superiority of mixing arbitrary images over existing label-specific and image-specific synthesis methods.

\subsection{Robustness Evaluation}\label{sec:robust}
Here, we evaluate the robustness of our method from two perspectives, that is, against unseen perturbations and against spatial deviations. Namely, perturbations may be easily applied to the images during data transportation, image compression, and bad light conditions of picture capturing. Spatial deviations are also common in real-world scenarios like unstable shooting positions or accidental data loss. All experiments for robustness are conducted on the FaceForensics++ dataset, and we take our backbone network Xception as the baseline.\\
\textbf{Robustness Against Unseen Perturbations.}
As shown in Fig.~\ref{fig:robust}, we introduce block-wise masking, median blur, and hue adjustment with multiple intensities for the robustness evaluation against unseen perturbations. To comprehensively investigate the robustness, we set the intensity of perturbations to gradually increase. Specifically, the perturbation intensities represent the masking ratios $\{0,0.05,0.10,0.15,0.20\}$ with fixed $8 \times 8$ grids for Block-wise, the kernel sizes $\{0,3,4,5,6\}$ for Blur, and hue adjustment ratios $\{0,0.1,0.2,0.3,0.4\}$ for Hue. Obviously, our method exhibits superior robustness due to the improved reception of forgery clues.\\
\textbf{Robustness Against Spatial Deviations.}
We provide a prediction stability study to investigate the robustness against spatial deviation. Specifically, we quantify the variation in prediction confidence for each image after applying the spatial deviations (\textit{i.e.}, Crop, Resize, and Shift) and subsequently depict these variations as a histogram in Fig.~\ref{fig:robust-spa}. Mathematically, the deviation value at each image can be written as $|\mathcal{N}(\mathbf{I})-\mathcal{N}(\mathcal{T}_s(\mathbf{I}))|$, where $\mathcal{N}(\cdot)$ is the detection network, $\mathbf{I}$ is the original image, and $\mathcal{T}_s$ is one of the spatial transforming operations.  A larger variation indicates a worse detection stability. For the baseline detector, a considerable number of detection results suffer from \reG{spatial deviations}. Especially for Crop, over one-third of the variations exceed 0.5, indicating a notable number of detection results are reversed to the opposite after spatial deviation. 
In contrast, with the help of AdvSCM, our detector can maintain strong robustness against different spatial deviations.

\begin{table}[tbp]
\centering
\caption{Results (AUC) of adding ED$^4$ to existing deepfake detection methods for plug-and-play effectiveness.} \label{tab:other}
\footnotesize
\centering
\begin{tabular}{lccc}\toprule
Method & DFD& CDFv2 &DFDC \\ \midrule
Effnb4                   &0.8004 & 0.7531          &   0.6848      \\
Effnb4 + ED$^4$   &\textbf{0.8550} \textcolor{red}{($\uparrow7\%$)}& \textbf{ 0.8129} \textcolor{red}{($\uparrow8\%$)}              &  \textbf{0.7214}  \textcolor{red}{($\uparrow5\%$)}  \\  \midrule
Capsule                   &0.7622 & 0.7062          &   0.6744      \\
Capsule + ED$^4$   &\textbf{0.8183} \textcolor{red}{($\uparrow7\%$)}& \textbf{ 0.7942} \textcolor{red}{($\uparrow12\%$)}              &  \textbf{0.7243}  \textcolor{red}{($\uparrow7\%$)}  \\  \midrule
SPSL                    &0.8151 & 0.7499         &   0.6859     \\
SPSL + ED$^4$   &\textbf{0.8656} \textcolor{red}{($\uparrow6\%$)}& \textbf{0.8115} \textcolor{red}{($\uparrow8\%$)}              &  \textbf{0.7361}  \textcolor{red}{($\uparrow7\%$)}\\ \bottomrule
\end{tabular}
\end{table}
\begin{figure*}[htbp]
    \centering
    \includegraphics[width=\linewidth]{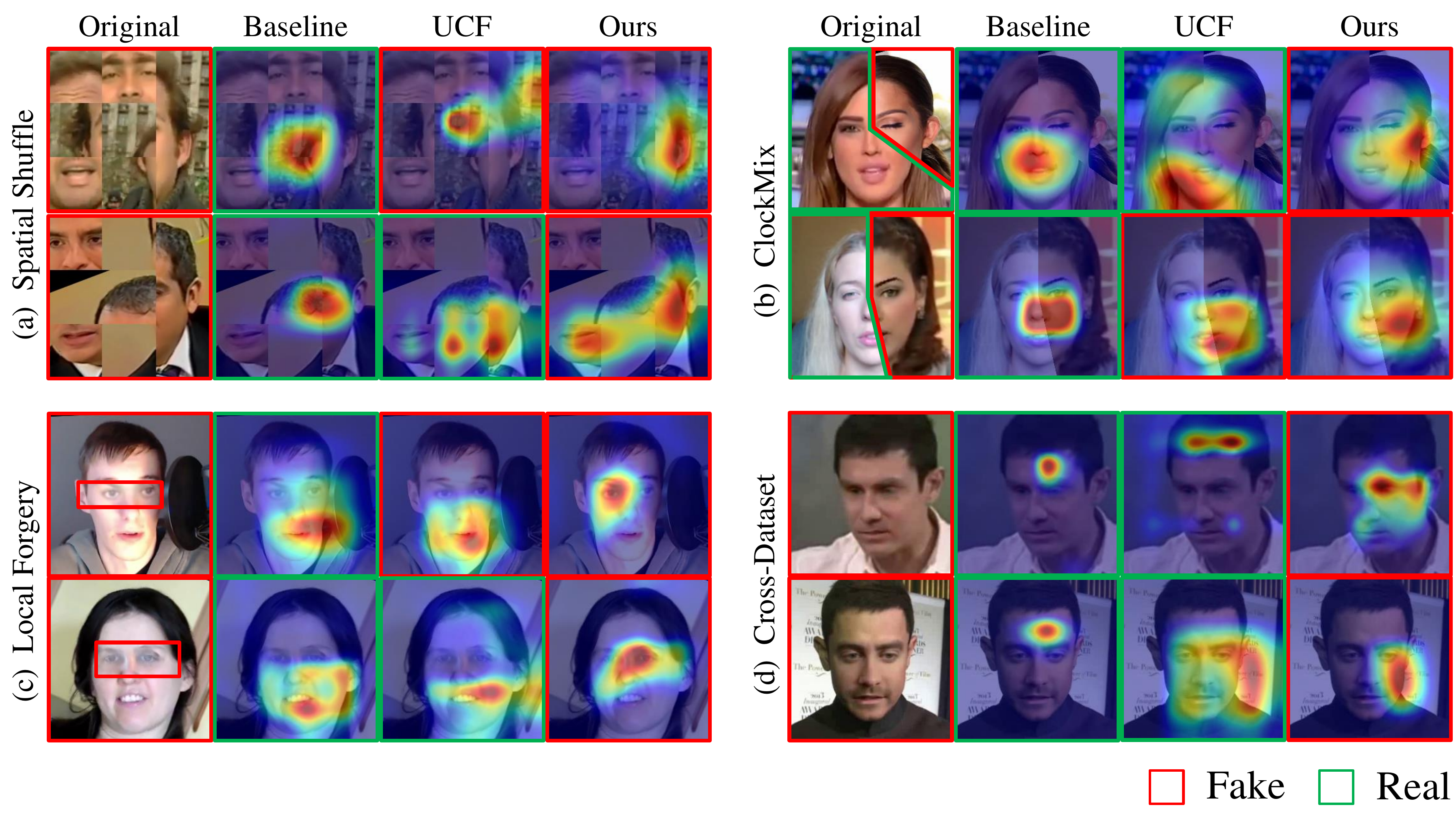}
    \caption{\reT{Class Activation Mapping (CAM)-based attention heatmaps for different methods on representative forgery images. ED$^4$ produces more concentrated and accurate attention on manipulated regions across diverse challenging scenarios, providing visual evidence of its superior sensitivity and localization precision.}}
    \label{fig:GradCam}
\end{figure*}
\subsection{Plug-and-Play with Previous Deepfake Detectors}
Considering our debiasing method is explicitly applied at the data level, it can be directly implemented to existing SOTA methods for deepfake detection. Therefore, we reproduce EfficientNetB4 (Effnb4)~\cite{effnet}, SPSL~\cite{spsl}, and Capsule~\cite{capsule} and then adding ED$^4$ to them. Tab.~\ref{tab:other} shows that with the additive implementation of ED$^4$, all detectors exhibit significant enhancements in performance. This substantiates that our method is plug-and-play that can be conveniently applied to other methods to enhance their effectiveness.

\subsection{Qualitative Results}
In this section, we visualize the Class Activation Map (CAM) via Grad-CAM++~\cite{grad-cam++} to discuss the focal regions of different detectors. UCF~\cite{ucf} is the most advanced method that addresses model bias by feature disentanglement via implicit network design. In Fig.~\ref{fig:GradCam}, we provide the CAM of images with four different conditions. Specifically, Spatial Shuffle (Fig.~\ref{fig:GradCam} (a)) and ClockMix (Fig.~\ref{fig:GradCam} (b)) are operations introduced by our method. Local Forgery (Fig.~\ref{fig:GradCam} (c)) is achieved by 
replacing eyes in the real faces with the fake ones following \cite{adv}. Images in Cross-Dataset (Fig.~\ref{fig:GradCam} (d)) are from CDFv1 while detectors are trained on FF++. 

Suffering from spatial bias, the Baseline is inertly focusing on the relatively central regions despite the facial regions being relocated or locally replaced. UCF can perceive forgery clues in a wilder region of the image center (the 2nd rows in Fig.~\ref{fig:GradCam} (a) and (c)), but it also struggles to precisely locate the forgery artifacts (Fig.~\ref{fig:GradCam} (b) and the 1st rows in Fig.~\ref{fig:GradCam} (a) and (c)). Conversely, our method can adaptively focus on the local forgery artifacts at diverse image locations.
Fig.~\ref{fig:GradCam} (d) indicates the challenges for the existing methods to detect the common artifacts in cross-dataset settings. Namely, the Baseline fails to perceive forgery, while UCF may overlook the actual forgery regions (the 2nd row in Fig.~\ref{fig:GradCam} (d)). Conversely, our method exhibits superior sensitivity to the cross-dataset common forgery artifacts. \reT{It should be noted that ED$^4$ is not designed to explicitly localize the precise forgery regions, the capability shown in Fig.~\ref{fig:GradCam} actually emerges from the following characteristics of ED$^4$:
\begin{itemize}
    \item The proposed ED$^4$ introduces strong data-level augmentation that effectively disrupts spurious correlation-based shortcuts for learning. As a result, the model has to become more sensitive to forgery artifacts and is compelled, during training, to implicitly learn to distinguish forged regions in order to produce correct predictions.
    \item During training, ED$^4$ is not applied to all inputs indiscriminately; instead, it is introduced at a controlled ratio to enrich the diversity of training samples. Consequently, the model implicitly acquires the ability to more accurately attend to and localize the forged regions.
\end{itemize}}

\reT{
\section{Limitations and Future Work}
\noindent \textbf{Limitation.} ED$^4$ is an explicit debiasing framework specifically designed for facial forgery detection. Accordingly, the proposed ClockMix requires pre-extracted facial landmarks to perform face alignment, ensuring a semantic consistency of facial regions during the mixing process. However, in broader AI-generated content (AIGC) detection scenarios, it is often infeasible to align semantically diverse images prior to applying ClockMix. As a result, ClockMix cannot be readily extended to general-purpose AIGC detectors and struggles to accommodate more diverse and unconstrained forgery scenarios.\\
\noindent \textbf{Future Work.} As semantic bias also poses a significant challenge in AIGC detection, the development of an explicit debiasing module tailored to this task remains an important direction for future research. One potential approach is to exploit the attention regions of the classifier to identify semantically salient areas within images. Semantic alignment could then be performed based on these regions, serving as a foundation for designing an explicit debiasing strategy adapted for AIGC detection.
}

\section{Conclusion}
In this paper, we improve the deepfake detector's generalizability from the debiasing perspective.
Firstly, besides content and specific-forgery biases, we reveal a new form of model bias termed spatial bias. That is, the deepfake detector consistently anticipates observing structural forgery clues at the image center. Then, we propose Explicit Data-level Debiasing for Deepfake Detection (ED$^4$), which is a unified framework to address the aforementioned three biases. Specifically, we propose ClockMix with multiple arbitrary faces to eliminate content and specific-forgery biases. For the spatial bias, we introduce the Adversarial Spatial Consistency Module (AdvSCM) that forces the backbone to extract spatial-agnostic features. Considering the improved effectiveness and plug-and-play convenience, ED$^4$ can be treated as a superior alternative to implicit disengagement by network design. Experiments demonstrate the promising performance of our method.
%
%
{
\small
\bibliographystyle{plain}
\bibliography{refer}
}
\end{document}